\documentclass[sigconf]{acmart}
\usepackage{multirow}
\usepackage{bm}
\usepackage{pifont}
\usepackage{subfigure}
\usepackage{pdflscape}
\usepackage{balance}
\usepackage{pgfplots}
\usepackage{makecell}

\newcommand{\VocTableSpacing}{1.015}
\newcommand{\CocoTableSpacing}{0.943}
\newcommand{\AblationTableSpacing}{1.00}
\newcommand{\VocTableFontsize}{\small}
\newcommand{\CocoTableFontsize}{\small}
\newcommand{\AblationTableFontSize}{\small}
\newcommand{\tablecaptionskip}{\setlength{\abovecaptionskip}{5pt}\setlength{\belowcaptionskip}{-2.5pt}}
\newcommand{\imagecaptionskip}{\setlength{\abovecaptionskip}{5pt}\setlength{\belowcaptionskip}{-2.5pt}}
\newcommand{\equationskip}{\setlength\abovedisplayskip{5pt}\setlength\belowdisplayskip{5pt}}
\newcommand{\sectionvspace}{}
\newcommand{\subsectionvspace}{\vspace{-1.7pt}}

\AtBeginDocument{%
	\providecommand\BibTeX{{%
			\normalfont B\kern-0.5em{\scshape i\kern-0.25em b}\kern-0.8em\TeX}}}

\copyrightyear{2022}
\acmYear{2022}
\setcopyright{none}
\makeatletter
\renewcommand\@formatdoi[1]{\ignorespaces}
\makeatother
\renewcommand{\footnotetextcopyrightpermission}[1]{}
\acmConference[MM '22]{Proceedings of the 30th
	ACM International Conference on Multimedia}{October 10--14, 2022}{Lisboa,
	Portugal}
\acmBooktitle{Proceedings of the 30th ACM International Conference on
	Multimedia (MM '22), October 10--14, 2022, Lisboa, Portugal}
\acmPrice{15.00}
\acmDOI{10.1145/3503161.3547834}
\acmISBN{978-1-4503-9203-7/22/10}

\acmSubmissionID{376}

\begin{document}
	
	\title{Global Meets Local: Effective Multi-Label Image Classification via Category-Aware Weak Supervision}
	
	
	\author{Jiawei Zhan}
	\authornote{indicates equal contributions.}
	\orcid{0000-0002-1745-4768}
	\author{Jun Liu}
	\authornotemark[1]
	\orcid{0000-0002-6985-8238}
	\affiliation{ 
		\institution{Tencent Youtu Lab, China}
		\country{}
	}
	
	\author{Wei Tang}
	\authornotemark[1]
	\affiliation{
		\institution{Institute of Automation, Chinese Academy of Sciences, China}
		\country{}
	}
	
	\author{Guannan Jiang}
	\author{Xi Wang}
	\affiliation{ 
		\institution{Contemporary Amperex Technology Co., Limited, China}
		\country{}
	}
	
	\author{Bin-Bin Gao}
	\orcid{0000-0003-2572-8156}
	\author{Tianliang Zhang}
	\orcid{0000-0002-3524-0878}
	\author{Wenlong Wu}
	\orcid{0000-0002-3732-6456}
	\affiliation{%
		\institution{Tencent Youtu Lab, China}
		\country{}
	}
	
	\author{Wei Zhang}
	\affiliation{%
		\institution{Contemporary Amperex Technology Co., Limited, China}
		\country{}
	}
	
	\author{Chengjie Wang}
	\authornote{indicates co-corresponding authors.}
	\affiliation{%
		\institution{Tencent Youtu Lab, China}
		\country{}
	}
	
	\author{Yuan Xie}
	\authornotemark[2]
	\affiliation{%
		\institution{East China Normal University, China}
		\country{}
	}
	
	\renewcommand{\shortauthors}{Jiawei Zhan et al.}
	%% No italics and no comma
	%% If needed use a foot or author note to identify equal contribution
	
	\begin{abstract}
		Multi-label image classification, which can be categorized into label-dependency and region-based methods, is a challenging problem due to the complex underlying object layouts. Although region-based methods are less likely to encounter issues with model generalizability than label-dependency methods, they often generate hundreds of meaningless or noisy proposals with non-discriminative information, and the contextual dependency among the localized regions is often ignored or over-simplified. This paper builds a unified framework to perform effective noisy-proposal suppression and to interact between global and local features for robust feature learning. Specifically, we propose category-aware weak supervision to concentrate on non-existent categories so as to provide deterministic information for local feature learning, restricting the local branch to focus on more high-quality regions of interest. Moreover, we develop a cross-granularity attention module to explore the complementary information between global and local features, which can build the high-order feature correlation containing not only global-to-local, but also local-to-local relations. Both advantages guarantee a boost in the performance of the whole network. Extensive experiments on two large-scale datasets (MS-COCO and VOC 2007) demonstrate that our framework achieves superior performance over state-of-the-art methods.
	\end{abstract}
	
	
	\maketitle
	\sectionvspace
	\section{Introduction}
	Multi-label image classification \cite{liu2020emerging,ganda2018survey,tidake2018multi}, which aims at predicting multiple labels for an image, is a fundamental task in the field of computer vision and multimedia. It has a wide range of applications in fields such as image retrieval \cite{chen2021deep}, attribute recognition \cite{thom2020facial}, and automatic image annotation \cite{wang2009multi}. Compared with single-label image classification, multi-label classification is more complex and challenging, as it is deals with complex underlying object layouts such as variations in location and scale and the difference between intra-class and inter-class.
	
	As one of the widely adopted solutions, region-based multi-label learning first generates a large number of proposals using methods such as selective search \cite{uijlings2013selective}, edge box \cite{zitnick2014edge} and BING \cite{cheng2014bing}. Binary cross-entropy loss is then used for each proposal instead of common softmax loss. More recent methods have begun to adopt long short-term memory \cite{RDAR,RARL} or utilize strong supervision \cite{yang2016exploit} such as bounding boxes with RPN \cite{ren2015faster} to generate more accurate proposals.
	
	Although acceptable results have been achieved, region-based methods still face two limitations: they generate a large number of noisy region proposals \cite{wei2014cnn,wei2015hcp,liu2018multi}, and the relationship among region proposals has not been thoroughly explored \cite{RARL,MCAR}. Specifically, to achieve a high recall rate, region-based methods mainly produce proposals using object-detection techniques. These methods usually generate many noisy region proposals, which are not only computationally inefficient for multi-label learning, but also detrimental to the performance due to the background interference and inaccurate border of the proposals. Moreover, as some methods are multi-stage \cite{ResNet-SRN,RARL} and do not explore the correlation among labels, region-based methods usually lack a thorough understanding of the global information of an image, and the multi-label information is not effectively utilized to learn the semantic relationships between regions. If the relationship is well established, the regional proposal can be further constrained. Efficiently addressing the limitations described above in a unified framework is therefore a crucial step in boosting performance.
	
	To generate region proposals more effectively, we propose category-aware weak supervision, accompanied with energy-based region-of-interest (ROI) selection to suppress noisy region proposals jointly. Since the ground-truth label only provides an indication of an existing category, rather than the accurate location of instances, current region-based methods are more likely to generate undesirable region proposals. Thus, we designed a weakly supervised loss function by only considering the feature map activation of non-existent categories, which can suppress most of the background information from an image through a weakly supervised learning process, reducing the number of proposals from thousands to dozens. Moreover, unlike the traditional area-based selection schema, the energy-based criterion can efficiently generate meaningful region proposals, thus enhancing their quality.
	
	To establish the correlation between region proposals, we introduce a cross-granularity attention module, which can integrate feature information of instances at different granularities. The determination of how many categories are presented in an image, as distinct from instance localization, is a comprehensive global process, whereas region proposal is a local procedure. With the help of the cross-granularity attention module, our approach enables the construction of a high-order global-to-local interaction, which not only builds the connection between global features and local features (global-to-local), but also explores the association between region features (local-to-local) to improve representation capability and further suppress noisy region proposals. Both advantages guarantee a boost in the performance of the framework. 
	
	Overall, the major contributions of this work can be summarized as follows:
	\begin{itemize}
		\item A novel and unified framework for multi-label image classification tasks is proposed that includes category-aware weak supervision and a cross-granularity attention module, thus significanlty reducing the number of proposals and enabling the modeling of correlations among regions.
		\item Unlike traditional methods of supervision, which focus on the instances of presented categories, our proposed category-aware weak supervision concentrates on non-existent categories to provide deterministic information in the process of learning local features, thus effectively suppressing noisy region proposals.
		\item The cross-granularity attention module can not only capture global-to-local relationships, but also mine local-to-local correlations. To our knowledge, this is the first region-based method that can explore high-order global-to-local correlations among region proposals.
		\item Extensive experiments were conducted on two challenging large-scale public datasets (MS-COCO \cite{lin2014microsoft} and VOC 2007 \cite{everingham2015pascal}), and the results demonstrate that our framework achieves superior performance over state-of-the-art methods.
	\end{itemize}
	
	\begin{figure*}[t]
		\centering
		\vspace{-5pt}
		\includegraphics[width=0.98\linewidth]{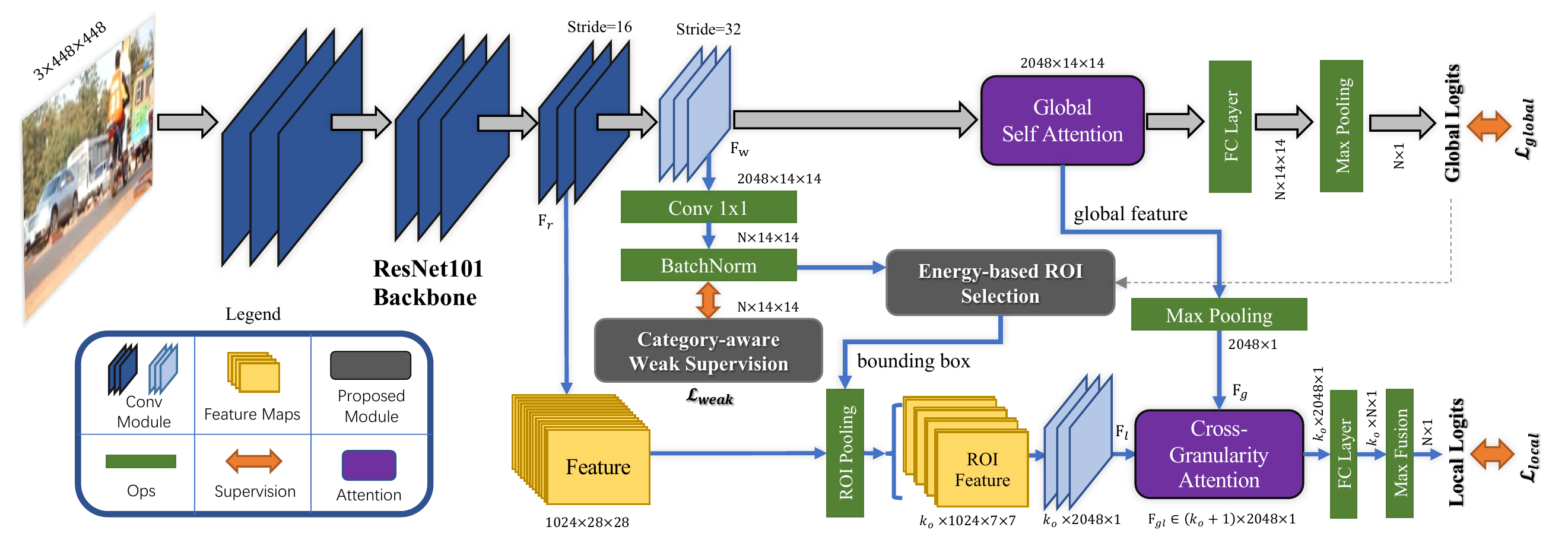}
		\imagecaptionskip
		\caption{The overview of our proposed framework.}
		\label{fig:overview}
	\end{figure*}
	
	\sectionvspace
	\section{Related work}
	
	Previous works on this task mainly follow two directions: label-dependency methods and region-based methods.
	
	\subsectionvspace
	\subsection{Label-dependency Methods}
	
	To solve the problem that the correlation among labels is usually ignored in multi-label classifications,
	label-dependency methods start with labels and establish correlation across categories through the co-occurrence between different labels.
	\cite{CNN-RNN} jointly learns image features and label correlations in a unified framework composed of a CNN module and an LSTM layer.
	Some works \cite{RDAR, RARL, RLSD} further took advantage of proposal generation/visual attention mechanism and LSTM to explicitly model label dependencies.
	But it requires an explicit module for removing duplicate prediction labels and needs a threshold for stopping the sequence outputs.
	
	Recently, due to the effectiveness of the Graph Convolutional Network in establishing correlations between labels,
	\cite{chen2019learning, chen2019multi, wang2020multi, ye2020attention} try to model the label dependency with a graph to boost the performance of multi-label image classification.
	While reasonably effective, these methods suffer from high computational costs or manually defined adjacency matrices. Besides, it is also arguable that it may learn spurious correlations when the label statistics are insufficient.
	
	As a result, since label-dependency methods rely on the prior of the training data to a great extent, these methods may degrade the model's generalizability when faced with domain shift.
	
	\subsectionvspace
	\subsection{Region-based Methods}
	
	Region-based methods are another hot branch for multi-label image classification.
	Most existing region-based methods focus on locating informative regions (e.g., proposal candidates \cite{wei2015hcp, yang2016exploit, RLSD}, attentive regions \cite{RDAR, DBLP:journals/corr/abs-2108-02456,MCAR}, random regions \cite{RCP}) to cover all possible existing objects and aggregate local discriminative features to facilitate recognizing multiple labels of the given image.
	
	Proposal candidates normally rely on object detection techniques \cite{girshick2015fast}, typically including BING \cite{wei2014cnn, cheng2014bing}, EdgeBox \cite{wei2015hcp, liu2018multi, zitnick2014edge} or Selective Search \cite{yang2016exploit, uijlings2013selective} to generate an arbitrary number of object segment hypotheses, which often obtain hundreds of meaningless or noisy proposals with non-discriminative information, and the modeling of spatial contextual dependency among localized regions is often ignored or over-simplified \cite{RARL,MCAR}.
	However, these methods model the spatial contextual dependency of different regions by using the category-agnostic mechanism, leading to generating noisy ROI proposals inevitably, as label knowledge is underutilization.
	
	Both \cite{yang2016exploit} and \cite{RLSD} utilize strong supervision (ground-truth bounding box annotations) to enhance the feature discriminative power. Despite the performance is improved, it will increase the cost of data acquisition.
	There have been some other attempts on multi-label researches, such as attention-based methods \cite{ResNet-SRN, ResNet101-ACfs, DBLP:journals/corr/abs-2108-02456}, dictionary learning \cite{DSDL} and few-shot multi-label classification \cite{chen2020knowledge}.
	
	In this paper, we aim to improve the performance of multi-label recognition with only image semantic information and propose category-aware weak supervision combined with an energy-based ROI selection schema to effectively suppress a large number of noisy region proposals. In addition, inspired by the methodology of label-dependency, by introducing cross-granularity attention, we establish global-to-local interaction to build the relationship not only between global and local features but also within different categories in an implicit way.
	\sectionvspace
	\section{Methodology}
	\subsectionvspace
	\subsection{Framework Overview}
	The framework consists of two branches: a global branch and a local branch. The global branch is used to generate the basic prediction for multi-label classification, and the local branch aims to provide category-specific (not only one category, but several saliency categories) prediction under category-aware weak supervision and energy-based ROI selection. The two branches interact with each other with the help of the cross-granularity attention module. Fig.\,\ref{fig:overview} provides an overview of our proposed framework.
	
	In the global branch, the samples are augmented and first passed through the backbone to produce feature maps with size $\!\frac{1}{16}\!$ and $\!\frac{1}{32}$, denoted by $\!\bm{F_{16}}\!$ and $\!\bm{F_{32}}$, respectively. The $\!\bm{F_{32}}\!$ features are processed through the global self-attention module, the fully connected layer, and a max-pooling layer to produce the prediction. Binary cross-entropy is then used as the objective function to train this branch:
	\begin{equation}\label{global-loss}
		\equationskip
		\mathcal{L}_{global}\!=\!-\sum_{j}^{L} (y_{j}\log{ (\overline{y}_{j}) }+(1-y_{j}) \log{ (1-\overline{y}_{j}) }),
	\end{equation}
	where $y_{j} \in \{0,1\}^L$ is the ground-truth of the j-th category of the image, which is a multi-hot binary vector with a set of $L$ labels in total. $\overline{y}_{j}$ is the prediction of the global branch.
	
	In the local branch, the main idea is to extract high-quality region proposals while eliminating large-scale noisy ones by using the proposed category-aware weak supervision and energy-based ROI selection mechanism. In detail, let $\bm{F_{w}}\in \mathbb{R}^{C_w\times H_w\times W_w}$ denote the feature maps that will be downsampled and used for weak supervision, where $H_w$, $W_w$ and $C_w$ are the height, width, and channels of the feature map, respectively. In our default experimental setup, we make $\bm{F_{w}}=\bm{F_{32}}$. 
	The $\bm{F_{w}}$ feature map is then connected to a 1$\!\times\!$1 convolution, reducing the number of channels to the number of categories $N$, and then connected to the batch normalization layer.
	
	After batch normalization, the pipeline splits into two modules: category-aware weak supervision and an energy-based ROI selection. 
	The weak supervision is used to improve feature representations of local regions and then suppress the noisy ROI proposals generated from non-existent categories. 
	The energy-based ROI selection allows for the automatic selection of ROIs based on the criterion of energy value rather than area size, which further filters noisy ROIs. 
	Next, the box coordinates generated by the energy-based ROI selection are fed into the ROI pooling layer to crop $\bm{F_{r}}$ and produce the output as local features. Here we have $\bm{F_{r}}=\bm{F_{16}}$.
	
	Finally, by taking as input $\bm{F_{g}}$ (global feature) and $\bm{F_{l}}$ (local feature associated with several categories), the proposed cross-granularity attention module can effectively explore the complementary information between global and local regions, making possible high-order interaction, including global-to-local and local-to-local (see the bottom right of Fig.\,\ref{fig:overview}).
	
	\subsectionvspace
	\subsection{Category-aware Weak Supervision}
	\label{sec:weakly}
	This module aims to improve feature representations of local regions by introducing auxiliary weak supervision. Since the dimensions of the feature maps have been reduced to the number of categories $N$, each feature map can be considered as an indicator of whether the corresponding category is active. In other words, the activation value of feature maps corresponding to non-existent categories should be suppressed, and, conversely, feature maps of existing categories should achieve a relatively high activation in the region near an instance.
	
\begin{figure}[t]
		\centering
		\includegraphics[width=\linewidth]{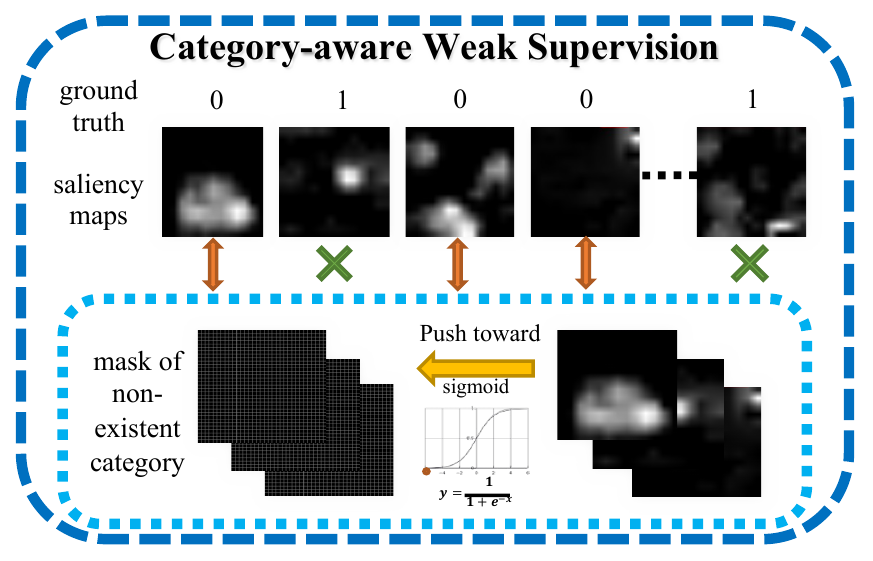}
		\imagecaptionskip
		\caption{The Category-aware Weak Supervision.}
		\label{fig:weakly}
	\end{figure}

	Since we are unable to accurately determine the location of instances at this stage, forcing a restriction on feature map activation tends to mislead the model into a sub-optimal situation. In any case, however, feature maps of non-existent categories should always be inactive. Thus, we designed a weakly supervised loss function that considers only feature maps of non-existent categories: when some categories do not exist, we expect the activation value of each pixel in their corresponding feature maps to be 0, as illustrated in Fig.\,\ref{fig:weakly}. Otherwise, we do not compute the loss for existing categories.
	\begin{equation}\label{weak-loss}
		\equationskip
		\mathcal{L}_{weak}=-\frac{1}{H\cdot W}\sum_{i=1}^{H}\sum_{j=1}^{W}{\frac{1}{N}\sum_{k\in\lnot gt}\!log\left (1-S ({\hat{p}}_{ij}^{k}) +\delta\right) }.
	\end{equation}
	This is a typical binary classification loss, where H and W are the height and width of the image, respectively, ${\hat{p}}_{ij}^{k}$ is the activation value of the corresponding pixel on the $k$-th category feature map, $S(\cdot)$ represents the sigmoid function, $gt$ stands for the set containing existing categories, $N$ indicates the number of non-existent categories, and $\delta$ is a small value to prevent math domain errors.
	
	The weakly supervised loss function focuses on filtering noisy activation regions that correspond to non-existent categories, which creates an inaccurate estimation of the foreground. Thus, fewer proposals are generated, and the selection of regions is more precise than those without weak supervision.
	
	\subsectionvspace
	\subsection{Energy-based ROI Selection}
	\label{sec:roi_select}
	
	The primary purpose of energy-based ROI selection is to extract the bounding box of the local region so that the classification accuracy can be improved when the features corresponding to the ROI are jointly classified again. For this purpose, we need regions with positive activation values as candidates, which indicate a higher probability of the region containing the target.
	Specifically, the batch normalization can be regarded as an operation that automatically learns local region thresholds, while the ReLU can be seen as a filter that eliminates negative feature values so that feature maps with only positive values can be obtained.
	We then sort the classification scores (the logit from the global branch prediction) in descending order and select ROI candidates for each selected feature map in the top $k_s$ feature maps.
	
	A common strategy for selecting ROIs is to choose regions with large ambiguous areas corresponding to connected regions \cite{RLSD,zhou2016learning} but small overall activation values. This strategy, however, leads to the incorrect area and potential performance degradation. The region with the higher overall activation is more likely to be the target we need. To this end, we chose to select the key ROIs by comparing the energy values $\mathbb{E}$ within the regions rather than by the region size: $\mathbb{E}\!=\!\sum_{i\in \mathcal{R}} \hat{p}_{i}$, where ${\hat{p}}_{i}$ is the $i$-th pixel activation within a region $\mathcal{R}$, 
	% which represents the smallest rectangle tangent to the boundary of the corresponding polygon of the activated region.
	which represents the minimum enclosing rectangle of the activated region boundary.
	
\begin{figure}[t]
		\centering
		\includegraphics[width=\linewidth]{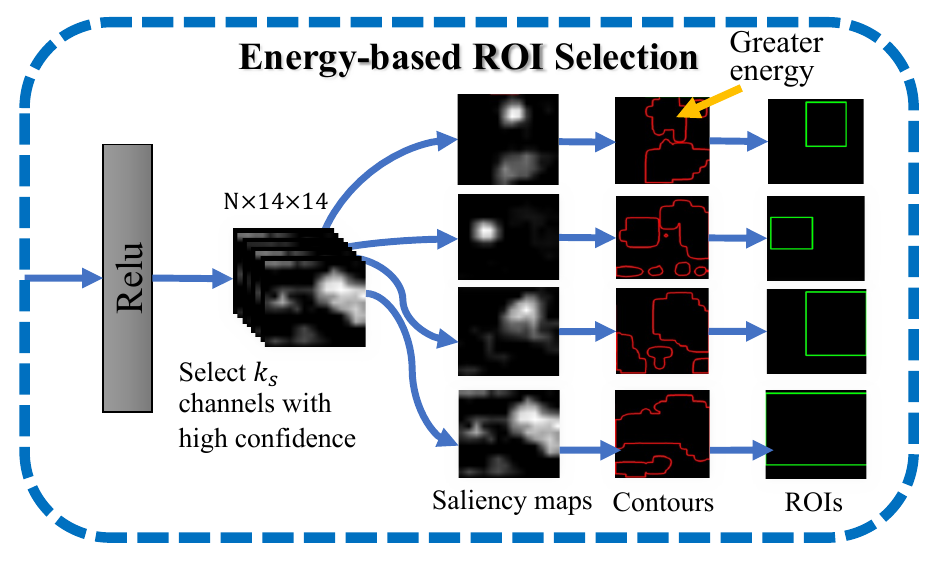}
		\imagecaptionskip
		\caption{The Energy-based ROI Selection.}
		\label{fig:selection}
	\end{figure}

	Once we obtain the energy value of each connected region in the feature map, we can generate high-quality ROIs by selecting the maximum energy and using the corresponding bounding boxes. As illustrated in Fig.\,\ref{fig:selection}, by treating $k_s$ feature maps as hints of objectness, 
	we find all the contours in each feature map and take the rectangular region of the contours with greater energy as the candidate. Note that we use the top $k_e$ region of the energy value to improve the coverage for the relevant target instance, and obtain $k_r$ bounding boxes of different sizes in the candidate energy region (further described in the supplementary material A since they are implementation details or tricks). In the end, we can obtain $k_o$ candidate regions $\mathcal{R}$, where $k_o = k_s\times k_r\times k_e$.
	
	\begin{table*}[t]
		\renewcommand\arraystretch{\VocTableSpacing}
		\centering
		\tablecaptionskip
		\caption{Comparisons of AP and mAP in $\%$ with state-of-the-art methods on the Pascal VOC 2007 dataset. 
			Input denotes the input image size during the inference. 
			\emph{$*$ denotes that the model is pretrained on the MS-COCO dataset.} 
			The best results and sub-optimal results are highlighted in {\color{red}red} and {\color{blue}blue}, respectively. Best viewed in color.
		}
		\VocTableFontsize
		\setlength\tabcolsep{2pt}
		\resizebox{\textwidth}{!}{
			\begin{tabular}{l|cc|cccccccccccccccccccc|c}
				\hline
				\textbf{Method} & \textbf{Bone} & \textbf{Input} & {\textbf{aero}} & {\textbf{bike}} & {\textbf{bird}} & {\textbf{boat}} & {\textbf{bottle}} & {\textbf{bus}} & {\textbf{car}} & {\textbf{cat}} & {\textbf{chair}} & {\textbf{cow}} & {\textbf{table}} & {\textbf{dog}} & {\textbf{horse}} & {\textbf{moto}} & {\textbf{person}} & {\textbf{plant}} & {\textbf{sheep}} & {\textbf{sofa}} & {\textbf{train}} & {\textbf{tv}} & {\textbf{mAP}} \\ \hline  \hline
				CNN-RNN \cite{CNN-RNN} & vgg-16 & 224   & 96.7  & 83.1  & 94.2  & 92.8  & 61.2  & 82.1  & 89.1  & 94.2  & 64.2  & 64.2  & 70.0    & 92.4  & 91.7  & 84.2  & 93.7  & 59.8  & 93.2  & 75.3  & \color{red}{99.7} & 78.6  & 84.0 \\ %\hline
				RMIC \cite{RMIC} & vgg-16 & 224   & 97.1  & 91.3  & 94.2  & 57.1  & {86.7}  & 90.7  & 93.1  & 63.3  & \color{blue}{83.3} & 83.3  & {92.8}  & 94.4  & 91.6  & 95.1  & 92.3  & 59.7  & 86.0    & 69.5  & 96.4  & 79.0    & 84.5 \\ %\hline
				RLSD \cite{RLSD} & vgg-16 & 224   & 96.4  & 92.7  & 93.8  & 94.1  & 71.2  & 92.5  & 94.2  & 95.7  & 74.3  & 74.3  & 74.2  & 95.4  & 96.2  & 92.1  & 97.9  & 66.9  & 93.5  & 73.7  & 97.5  & 87.6  & 88.5 \\ %\hline
				VeryDeep \cite{VeryDeep} & vgg-16 & 224   & 98.9  & 95.0    & 96.8  & 95.4  & 69.7  & 90.4  & 93.5  & 96.0    & 74.2  & 74.2  & 87.8  & 96.0    & 96.3  & 93.1  & 97.2  & 70.0    & 92.1  & 80.3  & 98.1  & 87.0    & 89.7 \\ %\hline
				ResNet101 \cite{ResNet}  & res-101 & 448 & 99.1  & 97.3  & 96.2  & 94.7  & 68.3  & 92.9  & 95.9  & 94.6  & 77.9  & 77.9  & 85.1  & 94.7  & 96.8  & 94.3  & 98.1  & 80.8  & 93.1  & 79.1  & 98.2  & 91.1  & 90.8 \\ %\hline
				HCP \cite{wei2015hcp}  & vgg-16 &    -   & 98.6  & 97.1  & 98.0    & 95.6  & 75.3  & 94.7  & 95.8  & 97.3  & 73.1  & 73.1  & 80.0    & 97.3  & 96.1  & 94.9  & 96.3  & 78.3  & 94.7  & 76.2  & 97.9  & 91.5  & 90.9 \\ %\hline
				RDAR \cite{RDAR}  & res-101 & 448  & 98.6  & 97.4  & 96.3  & 96.2  & 75.2  & 92.4  & 96.5  & 97.1  & 76.5  & 76.5  & 87.7  & 96.8  & 97.5  & 93.8  & 98.5  & 81.6  & 93.7  & 82.8  & 98.6  & 89.3  & 91.9 \\ %\hline
				FeV+LV \cite{yang2016exploit} & vgg-16 & 224   & 98.2  & 96.9  & 97.1  & 95.8  & 74.3  & 94.2  & 96.7  & 96.7  & 76.7  & 76.7  & 88.0    & 96.9  & 97.7  & 95.9  & 98.6  & 78.5  & 93.6  & 82.4  & 98.4  & 90.4  & 92.0 \\ %\hline
				RARL \cite{RARL} & res-101 & 448   & 98.6  & 97.1  & 97.1  & 95.5  & 75.6  & 92.8  & 96.8  & 97.3  & 78.3  & 78.3  & 87.6  & 96.9  & 96.5  & 93.6  & 98.5  & 81.6  & 93.1  & 83.2  & 98.5  & 89.3  & 92.0 \\ %\hline
				RCP  \cite{RCP} & vgg-16 &    -   & 99.3  & 97.6  & 98.0    & 96.4  & 79.3  & 93.8  & 96.6  & 97.1  & 78.0    & 78.0    & 87.1  & 97.1  & 96.3  & 95.4  & {99.1}  & 82.1  & 93.6  & 82.2  & 98.4  & 92.8  & 92.5 \\ %\hline
				ML-GCN \cite{chen2019multi} & res-101 & 448  & 99.5  & 98.5  & \color{red}{98.6}  & 98.1  & 80.8  & 94.6  & 97.2  & 98.2  & 82.3  & 82.3  & 86.4  & 98.2  & 98.4  & {96.7}  & \color{blue}{99.0}    & 84.7  & 96.7  & 84.3  & 98.9  & 93.7  & 94.0 \\ %\hline
				SSGRL \cite{chen2019learning}& res-101 & 448   & 99.5  & 97.1  & 97.6  & 97.8  & 82.6  & 94.8  & 96.7  & 98.1  & 78.0    & \color{blue}{97.0}    & 85.6  & 97.8  & 98.3  & {96.4}  & 98.8  &{84.9}  & 96.5  & 79.8  & 98.4  & 92.8  & 93.4 \\ %\hline
				A-GCN \cite{ye2020attention}& res-101 & 448   & 99.3 & 98.6 & 97.8 & 96.0 & 78.8 & 92.9  & 97.0 & 97.3  & 80.8  & 95.3  & 82.8 & 97.6 & \color{blue}{98.5} & 95.5 & 98.6 & 83.4 & \color{blue}{97.5} & 83.8 & 98.8 & 91.8 & 93.1 \\ %\hline
				CoP \cite{CoP}  & res-101 & 448   & \color{red}{99.9}  & 98.4  & 97.8  & {98.8}  & 81.2  & 93.7  & 97.1  &  98.4  & 82.7  & 94.6  & 87.1  & 98.1  & 97.6  & 96.2  & 98.8  & 83.2  & 96.2  & 84.7  & 99.1  & 93.5  & 93.8 \\ %\hline
				DSDL \cite{DSDL} & res-101 & 448   & \color{blue}{99.8}  & 98.7  & 98.4  & 97.9  & 81.9  & 95.4  & 97.6  & 98.3  & \color{blue}{83.3}  & 95.0    & 88.6  & 98.0    & 97.9  & 95.8  & \color{blue}{99.0}    & \color{blue}{86.6}  & 95.9  & 86.4  & 98.6  &  {94.4}  & 94.4 \\ %\hline
				MCAR \cite{MCAR} & res-101 & 448   & 99.7  & \color{blue}{99.0}  & \color{blue}{98.5}  & 98.2  & {85.4}  & {96.9}  & 97.4  & \color{blue}{98.9}  & \color{red}{83.7}  & {95.5}  & 88.8  & {99.1}    & 98.2  & 95.1  & \color{red}{99.1}    & 84.8  & {97.1}  & 87.8  & 98.3  & {94.8}  & 94.8 \\ \hline 
				Ours & res-101 & 448  &  \color{red}{99.9}  & {98.7} & 96.2 & \color{red}{99.7} & \color{blue}{85.7} & \color{blue}{97.3} & \color{blue}{98.3} & \color{red}{99.1}  & 76.0 & 95.2 & \color{red}{95.1} & \color{red}{99.4} & \color{red}{99.9} & \color{blue}{97.6} & \color{blue}{99.0} & 80.5 & {96.8} & \color{red}{93.7} & {99.2} & \color{blue}{96.6} & \color{blue}{95.2} \\ %\hline
				Ours$^*$ & res-101 & 448  & \color{red}{99.9}  & \color{red}{99.1} & 97.0 & \color{blue}{99.5} & \color{red}{89.1} & \color{red}{99.1} & \color{red}{99.5} & {98.4} & 81.0 & \color{red}{99.9} & \color{blue}{93.8} & \color{blue}{99.2} & \color{red}{99.9} & \color{red}{98.1} & \color{red}{99.1} & \color{red}{87.4} & \color{red}{99.2} & \color{blue}{92.5} & \color{blue}{99.5} & \color{red}{99.1}  & \color{red}{96.5} \\ \hline
			\end{tabular}%
		}
		\label{tab:overallvoc}%
	\end{table*}%
	\subsectionvspace
	\subsection{Cross-granularity Attention Module}
	The information spread between the global and local branches has usually been ignored in previous work. Since the global and local feature maps are misaligned in the spatial dimension, simple fusion does not provide a performance boost. We therefore resort to the self-attention mechanism to achieve global-to-local interaction.
	
	First, we perform self-attention on the global branch (see the purple rectangle at the top of Fig.\,\ref{fig:overview}) to capture non-local dependencies to generate a more high-level semantic feature. Let $Q_{g}\!=\!\phi^q_{g} ({\bm{F_{32}}}), K_{g}\!=\!\phi^k_{g} ({\bm{F_{32}}}), V_{g}\!=\!\phi^v_{g} ({\bm{F_{32}}}) $, where $\phi (\cdot) $ indicates a linear projection. The global self-attention maps can be calculated as
	\begin{equation}\label{global-att}
		\equationskip
		\bm{A_{g}}=Softmax(\frac{\bm{Q_{g}}^\mathsf{T}\bm{K_{g}}}{\sqrt{C}}), \bm{A_{g}}\in \mathbb{R}^{HW\times HW}.
	\end{equation}
	where the dot product is scaled by a factor $\frac{1}{\sqrt{C}}$ (where $C$ is the dimension of a query vector $Q_{g}$ and key vector $K_{g}$) to prevent its result from being too large \cite{attention}.
	
	A max-pooling operation is conducted on the optimized features $\bm{A_{g}}\bm{V_{g}}$ along the spatial dimension to obtain global features (i.e., $\bm{F_{g}}\!\in\!\mathbb{R}^{C\times 1}$), which are one of the inputs of the cross-granularity attention module.
	
	Another input of the cross-granularity attention module comes from the feature maps of selected ROIs (i.e., $\bm{F_{l}}\!\in\!\mathbb{R}^{k_o \times C\times 1}$), which is obtained by passing extracted features through a series of transform layers.
	By concatenating the features $\bm{F_{g}}$ and $\bm{F_{l}}$ at the batch dimension (i.e., $\bm{F_{gl}}\!\in\!\mathbb{R}^{ (k_o+1) \times (C\times 1) }$), we can calculate the self-attention to force information spreading between branches. Similar to Eq.\,(\ref{global-att}), let
	$\bm{Q_{gl}}\!=\!\phi^Q_{gl} ({\bm{F_{gl}}}), \bm{K_{gl}}\!=\!\phi^K_{gl} ({\bm{F_{gl}}}), \bm{\bm{V_{gl}}}\!=\!\phi^V_{gl} ({\bm{F_{gl}}}) $.
	The cross-granularity attention maps can be calculated as,
	\begin{equation}
		\equationskip
		\bm{A_{gl}}=Softmax(\bm{Q_{gl}}\bm{K_{gl}}^\mathsf{T}), \bm{A_{gl}}\in \mathbb{R}^{(k_o+1)\times (k_o+1)}.
	\end{equation}
	The output of the cross-granularity attention module, $\bm{A_{gl}}\bm{\bm{V_{gl}}}$, is then further sliced and summed as needed for the local branch and fed into the fully connected layer for classification to obtain local results.
	With the help of category-aware weak supervision and energy-based ROI selection, which restricts the local branch to concentrating on high-quality ROIs, noisy ROIs are significantly suppressed. These regions contain detailed information that the global branch cannot provide. By taking as inputs $\bm{F_{g}}$ and $\bm{F_{l}}$, the cross-granularity attention module can effectively explore the complementary information between global and local features. Moreover, self-attention allows the correlations to be captured within features of different categories, which build the label-dependency in an implicit manner.
	
	\subsectionvspace
	\subsection{Network Training}
	There are three kinds of losses combined to train the whole network. 
	\ding{172} $\mathcal{L}_{global}$ defined in Eq.\,(\ref{global-loss}). 
	\ding{173} $\mathcal{L}_{weak}$ to suppress noisy proposals of ROI, which are described in Eq.\,(\ref{weak-loss}).
	\ding{174} $\mathcal{L}_{local}$ is a cross entropy loss used to calculate the loss of local branches as follows,
	\begin{equation}\label{local-loss}
		\equationskip
		\mathcal{L}_{local}=-\sum_{j}^{L} (y_{j}\log{ (\widetilde{y}_{j}) }+ (1-y_{j}) \log{ (1-\widetilde{y}_{j}) }),
	\end{equation}
	where $y_{j} \in \{0,1\}^L$ is the ground-truth of the j-th category of the image, which is actually a multi-hot binary vector with a set of $L$ labels in total. $\overline{y}_{j}$ is the prediction after layers of cross-granularity attention, linear projection and max fusion in the local branch.
	
	Finally, the overall loss function is,
	\begin{equation}
		\equationskip
		\mathcal{L}_{total}=\mathcal{L}_{global}+\mathcal{L}_{weak}+\mathcal{L}_{local}.
	\end{equation}
	The proposed framework is trained by reducing $\mathcal{L}$ in an end-to-end fashion. 
	The final result is the average of the prediction results for both global and local branches.
	
	\sectionvspace
	\section{Experiment}
	In this section, we compare our proposed model with the state-of-the-art multi-label classification methods on two popular benchmark datasets: Pascal VOC 2007 \cite{everingham2015pascal} and Microsoft COCO \cite{lin2014microsoft}. In addition, comprehensive ablation and qualitative studies of the proposed method are also provided.
	
	\textbf{Implementation Details.}
	We adopt ResNet101 \cite{ResNet} as the backbone, which is pre-trained on ImageNet \cite{krizhevsky2012imagenet}.
	In the training phase, the image is first scaled to (N+64)$\times$(N+64), and then cropped at five scales [1.0, 0.875, 0.75, 0.66, 0.5] as suggested in \cite{chen2019multi} to avoid over-fitting. Finally, the cropped patches are further resized to N$\times$N. 
	During inference, the image is directly scaled to the size of N$\times$N, normalized for evaluation. 
	In addition, we set $k_s$=4, $k_r$=3, $k_e$=2 in our experiments.
	Our network was trained on 8 GPUs using the SGD algorithm with a total of 50 epochs, a batch size of 16, and initial learning rates of 0.05 and 0.0125 for VOC 2007 and MS-COCO, respectively, with a multi-step learning rate decay (i.e., [30, 40]).
	
	\textbf{Evaluation Metrics.}
	We employ the average precision (AP) for each category, and the mean average precision (mAP) overall categories to evaluate all the methods. In the experiments, we computed the overall precision, recall, F1 (OP, OR, OF1) and per-class precision, recall, F1 (CP, CR, CF1) for comparison.

	\begin{table*}[t]
		\renewcommand\arraystretch{\CocoTableSpacing}
		\centering
		\footnotesize
		\tablecaptionskip
		\caption{Comparisons with state-of-the-art methods on the MS-COCO dataset. 
			% The Input column represents the input image size for the inference phase.
			Input denotes the input image size during the inference. 
			\emph{$*$ denotes that the backbone is replaced with ResNeXt-101 32x8d pre-trained on ImageNet using semi-supervised methods.}
			The best results and sub-optimal results are highlighted in {\color{red}red} and {\color{blue}blue}, respectively. Best viewed in color.
		}
		\CocoTableFontsize
		\setlength\tabcolsep{6.5pt}
		\begin{tabular}{l|cc|c|cccccc|cccccc}
			\hline
			\multirow{2}{*}{\textbf{Methods}} & \multirow{2}{*}{\textbf{Bone}} &  \multirow{2}{*}{\textbf{Input}} & \multirow{2}{*}{\textbf{mAP}} & \multicolumn{6}{c|}{\textbf{ALL}} & \multicolumn{6}{c}{\textbf{TOP3}} \\ \cline{5-16}
			&&&& \textbf{CP} & \textbf{CR} & \textbf{CF1} & \textbf{OP} & \textbf{OR} & \textbf{OF1} & \textbf{CP} & \textbf{CR} & \textbf{CF1} & \textbf{OP} & \textbf{OR} & \textbf{OF1} \\ \hline \hline
			CNN-RNN \cite{CNN-RNN} & vgg-16 & 224   & -     & -     & -     & -     & -     & -     & -     & 66.0    & 55.6  & 60.4  & 69.2  & 66.4  & 67.8 \\ %\hline
			RLSD \cite{RLSD} & vgg-16 & 224   & -     & -     & -     & -     & -     & -     & -     & 67.6  & 57.2  & 62.0    & 70.1  & 63.4  & 66.5 \\ %\hline
			RDAR \cite{RDAR} & vgg-16 & 448   & -     & -     & -     & -     & -     & -     & -     & 79.1  & 58.7  & 67.4  & 84.0    & 63.0    & 72.0 \\ %\hline
			RARL \cite{RARL} & vgg-16 & 448   & -     & -     & -     & -     & -     & -     & -     & 78.8  & 57.2  & 66.2  & 84.0    & 61.6  & 71.1 \\ %\hline
			
			ResNet101 \cite{ResNet} & res-101 & 224   & 77.3  & 80.2  & 66.7  & 72.8  & 83.9 & 70.8  & 76.8  & 84.1  & 59.4  & 69.7 & 89.1  & 62.8  & 73.6 \\ %\hline
			SRN \cite{ResNet-SRN} & res-101 & 224   & 77.1  & 81.6  & 65.4  & 71.2  & 82.7  & 69.9  & 75.8  & 85.2  & 58.8  & 67.4  & 87.4  & 62.5  & 72.9 \\ %\hline
			Acfs \cite{ResNet101-ACfs} & res-101   & 224   & 77.5  & 77.4  & 68.3  & 72.2  & 79.8  & 73.1  & 76.3  & 85.2  & 59.4  & 68.0    & 86.6  & 63.3  & 73.1 \\ %\hline
			MultiEvid \cite{Multi-Evidence} & res-101 & 448   & -     & 80.4  & 70.2  & 74.9  & 85.2  & 72.5  & 78.4  & 84.5  & 62.2  & 70.6  & 89.1  & 64.3  & 74.7 \\ %\hline
			
			DecoupleNet \cite{DecoupleNet} & res-101 & 448   & 82.2  & 83.1  & 71.6  & 76.3  & 84.7  & 74.8  & 79.5  & -     & -     & -     & -     & -     & - \\ %\hline
			ML-GCN \cite{chen2019multi} & res-101  & 448   & 83.0    & \color{blue}{85.1}  & 72.0    & \color{blue}{78.0}    & 85.8  & 75.4  & \color{blue}{80.3}  & \color{blue}{89.2}  & 64.1  & 74.6  &  {90.5}  & \color{blue}{66.5}  & \color{blue}{76.7} \\ %\hline
			
			CoP \cite{CoP}  & res-101 & 448   & 81.1  & 81.2  & 70.8  & 75.8  & 83.6  & 73.3  & 78.1  & 86.4  & 62.9  & 72.7  & 88.7  & 65.1  & 75.1 \\ %\hline
			DSDL \cite{DSDL} & res-101 & 448   & 81.7  & 84.1  & 70.4  & 76.7  & 85.1  & 73.9  & 79.1  & 88.1  & 62.9  & 73.4  & 89.6  & 65.3  & 75.6 \\ %\hline
			CSRA \cite{DBLP:journals/corr/abs-2108-02456} & res-101  & 448   & 83.5    & 84.1  & \color{blue}{72.5}    & 77.9    & 85.6  & \color{blue}{75.7}  & \color{blue}{80.3}  &  {88.5}  & \color{blue}{64.2}  & 74.4  & 90.4  & {66.4}  & 76.5 \\ %\hline
			MCAR \cite{MCAR} & res-101 & 448   & \color{blue}{83.8} &  {85.0} & 72.1 & \color{blue}{78.0} & \color{red}{88.0} & 73.9 & \color{blue}{80.3} & 88.1 & \color{red}{65.5} & \color{blue}{75.1} &\color{blue}{91.0} & 66.3 & \color{blue}{76.7} \\ %\hline
			Ours & res-101 & 448   &  \color{red}{85.3} &  \color{red}{86.3}  &  \color{red}{74.4}  &  \color{red}{79.9}  &  \color{blue}{87.5}  &  \color{red}{77.6}  &  \color{red}{82.3}  & \color{red}{89.6}  &  \color{red}{65.5}  &  \color{red}{75.7} &  \color{red}{91.7}  &  \color{red}{67.9}  &  \color{red}{78.1} \\  \hline
			
			SSGRL \cite{chen2019learning} & res-101 & 576   & 83.8  & \color{red}{89.9}  & 68.5  & 76.8  & \color{red}{91.3}  & 70.8  & 79.7  & \color{red}{91.9}  & 62.5  & 72.7  & \color{red}{93.8}  & 64.1  & 76.2 \\ %\hline
			KGGR \cite{chen2020knowledge} & res-101 & 576 &  84.3 & 85.6  & {72.7}  & 78.6  & 87.1  & 75.6  & 80.9  & 89.4  & {64.6}  & 75.0  & 91.3  &  66.6  & {77.0} \\ %\hline
			MCAR \cite{MCAR} & res-101 & 576   & 84.5 & 84.3 & {73.9} & {78.7} & 86.9 & {76.1} &{81.1} & 87.8 & {65.9} &{75.3} & 90.4 & {67.1} & {77.0} \\ %\hline
			A-GCN \cite{ye2020attention} & res-101 & 576   & \color{blue}{85.2}    & 84.7  & \color{red}{75.9}   & \color{blue}{80.1}   & 84.9  & \color{red}{79.4}  & \color{blue}{82.0}  & 88.8  & \color{red}{66.2}  & \color{blue}{75.8}  & 90.3  & 68.5  & \color{blue}{77.9} \\ 
			C-Tran \cite{CTran} & res-101 & 576 & 85.1 & \color{blue}{86.3}  & \color{blue}{74.3}  & 79.9  & {87.7}  & 76.5  & 81.7  & \color{blue}{90.1}  & 65.7  & \color{red}{76.0}  & 92.1  & \color{red}{71.4}  & 77.6 \\
			Ours  & res-101 &  576   & \color{red}{86.7} & {86.1}  & \color{red}{75.9}  & \color{red}{80.6} & \color{blue}{88.5} & \color{blue}{78.9}  & \color{red}{83.4} & {89.0}  & \color{blue}{66.1}  & \color{blue}{75.8}   & \color{blue}{92.3} & \color{blue}{68.6}  & \color{red}{78.7} \\ \hline
			Ours*  & next-101 &  576   & \color{red}{88.8} & \color{red}{88.8} & \color{red}{75.9} & \color{red}{81.9} & \color{red}{88.9} & \color{red}{79.2} & \color{red}{83.8} & \color{red}{91.3} & \color{red}{66.1} & \color{red}{76.7} & \color{red}{92.3} & \color{red}{69.1} & \color{red}{79.0} \\ \hline
		\end{tabular}%
		\label{tab:overallcoco}%
	\end{table*}%
	
	\subsectionvspace
	\subsection{Comparisons with State-of-the-Arts} 
	\textbf{Result on VOC 2007.} The Pascal VOC 2007 dataset contains 9,963 images of 20 object categories, with about 1.5 labels per image. 
	We compare our method against the following SoTA methods: 
	CNN-RNN \cite{CNN-RNN}, 
	RMIC \cite{RMIC}, 
	RLSD \cite{RLSD}, 
	VeryDeep \cite{VeryDeep}, 
	ResNet101 \cite{ResNet}, 
	HCP \cite{wei2015hcp}, 
	RDAR \cite{RDAR}, 
	FeV+LV \cite{yang2016exploit}, 
	RARL \cite{RARL}, 
	RCP \cite{RCP}, 
	ML-GCN \cite{chen2019multi}, 
	SSGRL \cite{chen2019learning},
	A-GCN \cite{ye2020attention},
	CoP \cite{CoP}, 
	DSDL \cite{DSDL} and 
	MCAR \cite{MCAR}.

	As shown in Tab.\,\ref{tab:overallvoc}, with an input size of 448$\times$448, our method achieves 95.2\% mAP, \emph{a new state-of-the-art performance on VOC 2007 dataset}. It outperforms the previous methods such as SSGRL \cite{chen2019learning} by 1.8\%,
	CoP \cite{CoP} by 1.4\%, ML-GCN \cite{chen2019multi} by 1.2\%. Finally, our method achieves 96.5\% mAP with the COCO pretrained model.
	
	\textbf{Result on MS-COCO.} MS-COCO contains 122,218 images of 80 object labels, with about 2.9 labels per image. 
	We followed the official split of 82,081 images for training and 40,137 images for testing.
	We compare our method against the following SoTA methods:
	CNN-RNN \cite{CNN-RNN}, 
	RLSD \cite{RLSD}, 
	RDAR \cite{RDAR}, 
	RARL \cite{RARL},
	DELTA \cite{DELTA},
	ResNet101 \cite{ResNet}, 
	SRN \cite{ResNet-SRN}, 
	Acfs \cite{ResNet101-ACfs}, 
	MultiEvid \cite{Multi-Evidence}, 
	DecoupleNet \cite{DecoupleNet}, 
	ML-GCN \cite{chen2019multi}, 
	CoP \cite{CoP},
	DSDL \cite{DSDL}, 
	CSRA \cite{DBLP:journals/corr/abs-2108-02456}, 
	MCAR \cite{MCAR}, 
	SSGRL \cite{chen2019learning}, 
	KGGR \cite{chen2020knowledge},
	A-GCN \cite{ye2020attention} and
	C-Tran \cite{CTran}.
	
	As shown in Tab.\,\ref{tab:overallcoco}, our method achieves 85.3\% mAP, which outperforms CoP \cite{CoP} by 4.2\%, DSDL \cite{DSDL} by 3.6\%, ML-GCN \cite{chen2019multi} by 2.3\%, CSRA \cite{DBLP:journals/corr/abs-2108-02456} by 1.8\%, respectively.
	Finally, our method achieves 88.8\% mAP with a semi-weakly supervised \cite{swsl} ResNeXt-101 32x8d \cite{resnext} as the visual feature extractor, surpassing other methods including the transformer-based.
	The remarkable improvement owes to our proposed framework, which guarantees the generation of high-quality region proposals efficiently as well as the exploration of the complementary information between global and local.
	
	\subsectionvspace
	\subsection{Ablation Studies}
	
	\begin{figure}[t]
		\centering
		\includegraphics[width=\linewidth]{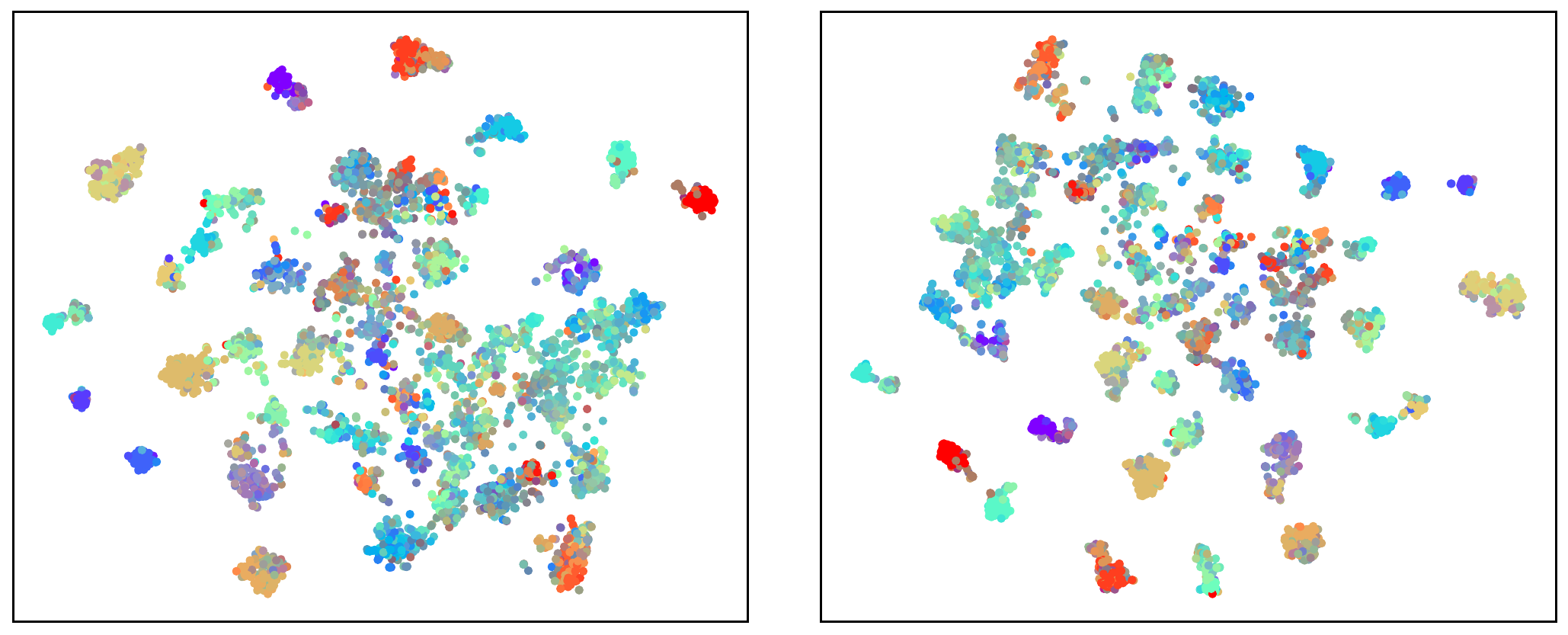}
		\imagecaptionskip
		\caption{t-SNE visualization for models trained without weak supervision (left) and with the combination of weak supervision (right). 
			Each dot represents one label-level embedding of a unique tag and each color represents one class.
		}
		\label{fig:weak_visual}
	\end{figure}
	
	To explain how our method works, we conducted exhaustive experiments to study the influence of different modules on the performance of multi-label image classification tasks.
	
	\textbf{Weak Supervision.}
	To demonstrate the effectiveness of our weak supervision, we show the t-SNE \cite{van2008visualizing} visualization of the label-level embeddings of samples on test sets of the MS-COCO dataset \cite{lin2014microsoft} in Fig.\,\ref{fig:weak_visual}. These embeddings are trained with weak supervision (i.e., Eq.\,(\ref{weak-loss})) and without weak supervision, respectively. From Fig\,\ref{fig:weak_visual}, additional weak supervision allows the embeddings of the same label to fall into more compact clusters, which are better separated from the clusters of other labels. This clearly shows the enforced distinctiveness of the embeddings with weak supervision.
	
	\textbf{Cross-granularity Attention.}
	To illustrate the effectiveness of our proposed attention, we tested three different variants separately on VOC 2007 and MS-COCO as shown in Tab.\,\ref{tab:attention}. 
	In Our w/o CA, $\bm{F_{l}}$ is directly used to obtain the classification results, and no $\bm{F_{g}}$ is involved in the process. In Our w/o CA\&GA, $\bm{F_{32}}$ is directly used for classification instead of feeding into the self-attention module in the global branch. 
	
	\begin{table}[t]
		\renewcommand\arraystretch{\AblationTableSpacing}
		\centering
		\tablecaptionskip
		\caption{
			Comparison of mAP (\%) of our model (Original Model), 
			our model without cross-granularity attention (Our w/o CA),
			our model without cross-granularity attention and global self-attention (Our w/o CA, GA).}
		\setlength\tabcolsep{5pt}
		\AblationTableFontSize
		\begin{tabular}{l|c|cc|c|cc}
			\hline
			\multirow{2}{*}{\textbf{Methods}} & \multicolumn{3}{c}{\textbf{VOC 2007}} & \multicolumn{3}{|c}{\textbf{MS-COCO}}\\ \cline{2-7}
			& \textbf{mAP} & \textbf{CF1} & \textbf{OF1} & \textbf{mAP} & \textbf{CF1} & \textbf{OF1} \\
			\hline \hline
			Original Model & 95.2 & 89.0 & 90.7 & 85.3 & 79.1 & 82.2\\
			Our w/o CA & 94.5 & 88.1 & 90.2 & 83.3 & 77.0 & 80.3\\
			Our w/o CA\&GA & 94.1 & 87.3 & 89.2 & 81.7 & 75.1 & 78.7\\ \hline
		\end{tabular}%
		% }
	\label{tab:attention}%
\end{table}%

Cross-granularity attention can effectively explore the complementary information between branches, thus enabling global to local interaction. 
In addition, inter-pixel correlations can also be captured due to global self-attention. 
Consequently, both advantages guarantee a boost in the performance, which verified the effectiveness of both modules.

\textbf{Number of Feature Maps $k_s$ Chosen for ROI Selection.} 
As previously described, our proposed method needs fewer region proposals to achieve better performance. We choose the feature maps ranking top-$k_s$ scores from the global branch to find region proposals. 
Tab.\,\ref{tab:topK} shows the performance of our model under the different $k_s$ selected. The base method represents the result of no region selected. 
Since the VOC dataset contains an average of 1.5 categories per image and a maximum of 6 categories and the COCO dataset contains an average of 3.5 categories per image, increasing $k_s$ will result in more redundant non-existence categories, which will distract the model's attention and degrade its performance. 
From Tab.\,\ref{tab:topK}, our model achieves the best result with $k_s$=4. This proves that our method for region generation is very efficient.

\begin{table}[ht]
	\renewcommand\arraystretch{\AblationTableSpacing}
	\centering
	\tablecaptionskip
	\caption{Ablation study on the impact of the number of high-scoring heat maps on the performance. Experiments are conducted on the COCO dataset.}
	\setlength\tabcolsep{6.5pt}
	\AblationTableFontSize
	\begin{tabular}{l|l|llllll}
		\hline
		\multirow{2}{*}{\textbf{Num.}} & \multirow{2}{*}{\textbf{mAP}} & \multicolumn{6}{c}{\textbf{ALL}} \\ \cline{3-8}
		&& \textbf{CP} & \textbf{CR} & \textbf{CF1} & \textbf{OP} & \textbf{OR} & \textbf{OF1} \\ \hline \hline
		Base  & 83.2  & 84.0  & 71.2  & 77.1  & 86.6  & 75.2  & 80.5   \\ %\hline
		Top2  & 84.1  & 83.8  & 73.5  & 78.3  & 86.1  & 77.1  & 81.3   \\ %\hline
		Top4  & 85.3  & 86.3  & 74.4  & 79.9  & 87.6  & 77.6  & 82.3   \\ %\hline
		Top8  & 84.7  & 86.4  & 73.6  & 79.5  & 86.4  & 76.7  & 81.3   \\ \hline
	\end{tabular}%
	\label{tab:topK}%
\end{table}%

\textbf{Training Loss from Different branches.} Our whole network has three kinds of training loss: $\mathcal{L}_{global}$, $\mathcal{L}_{weak}$, and $\mathcal{L}_{local}$. Tab.\,\ref{tab:loss} shows the results of different combinations of losses. 
The mAP on both datasets improves 1.9\% and 2.1\%, respectively, when utilizing the category-aware weak supervision and the energy-based ROI selection.

\begin{table}[ht]
	\renewcommand\arraystretch{\AblationTableSpacing}
	\centering
	\tablecaptionskip
	\caption{Ablation study on the impact of the loss on the performance. Experiments are conducted on both VOC and COCO dataset.}
	\setlength\tabcolsep{10.5pt}
	\AblationTableFontSize
	\begin{tabular}{l|c|c}
		\hline
		\textbf{Training Loss}&\textbf{VOC}&\textbf{COCO} \\
		\hline \hline
		$\mathcal{L}_{global}$ & 93.21 & 83.25 \\
		$\mathcal{L}_{global}+\mathcal{L}_{local}$ & 94.27 & 84.60 \\
		$\mathcal{L}_{global}+\mathcal{L}_{local}+\mathcal{L}_{weak}$ & 95.17 & 85.31 \\ \hline
	\end{tabular}%
	
	\label{tab:loss}%
\end{table}%

\begin{figure*}[!t]
	\centering
	\vspace{5pt}
	\begin{minipage}[b]{1\textwidth}
		\includegraphics[width=\linewidth]{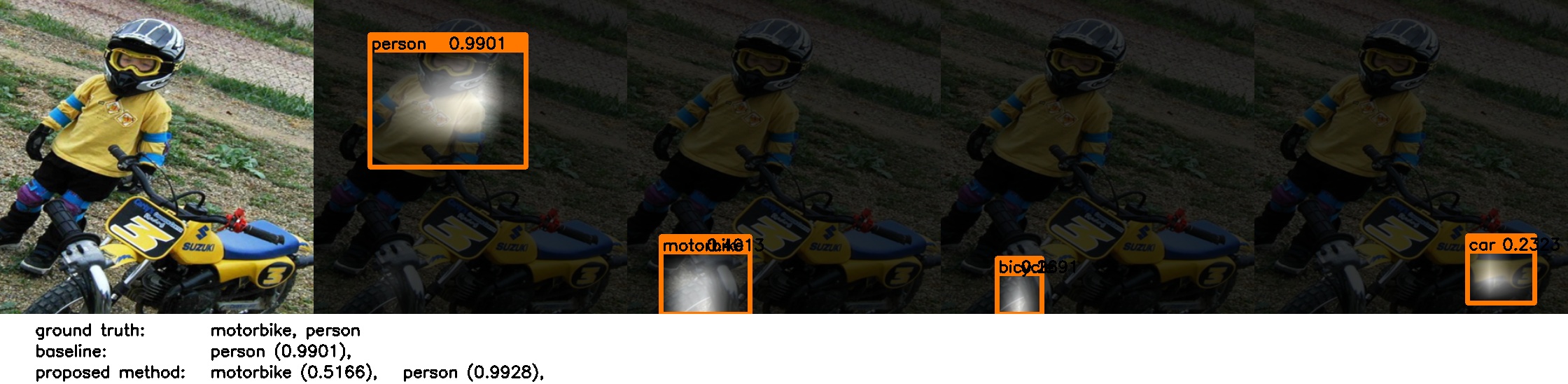}
		\vspace{-25pt}
		\begin{center}(a)\end{center}
		\vspace{5pt}
	\end{minipage}
	\begin{minipage}[b]{1\textwidth}
		\includegraphics[width=\linewidth]{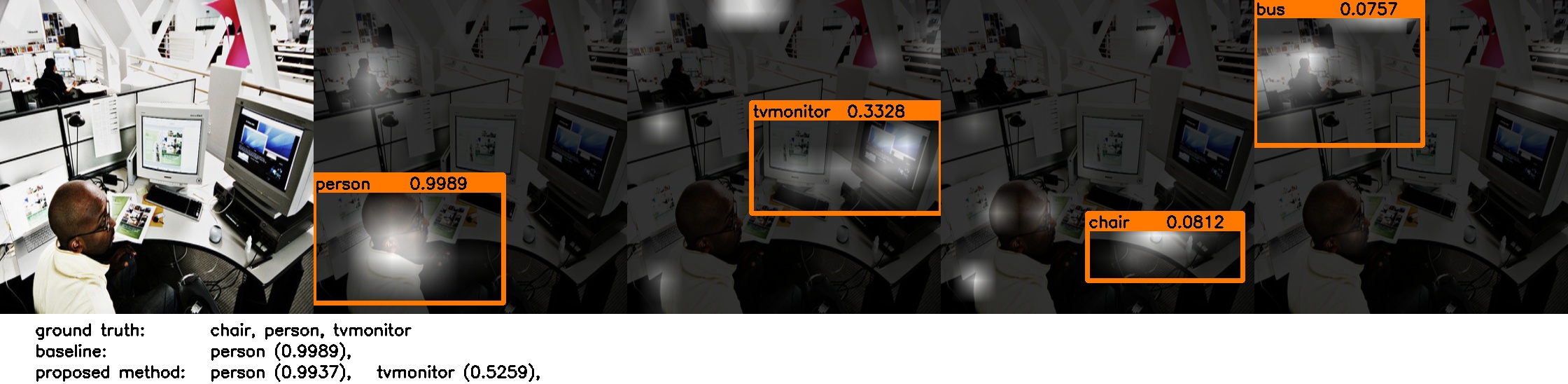}
		\vspace{-25pt}
		\begin{center}(b)\end{center}
		\vspace{5pt}
	\end{minipage}
	\begin{minipage}[b]{1\textwidth}
		\includegraphics[width=\linewidth]{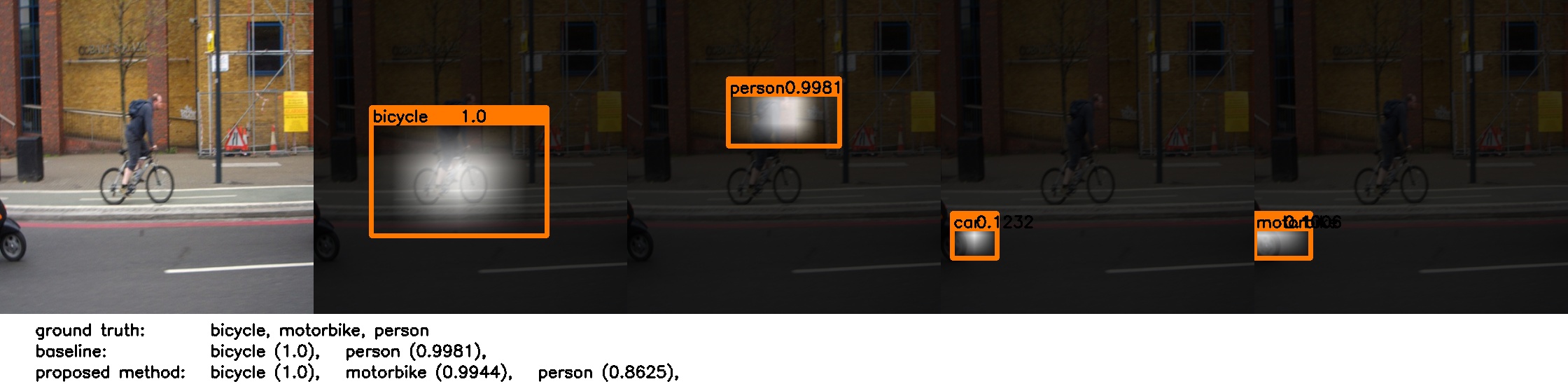}
		\vspace{-25pt}
		\begin{center}(c)\end{center}
		\vspace{5pt}
	\end{minipage}
	\begin{minipage}[b]{1\textwidth}
		\includegraphics[width=\linewidth]{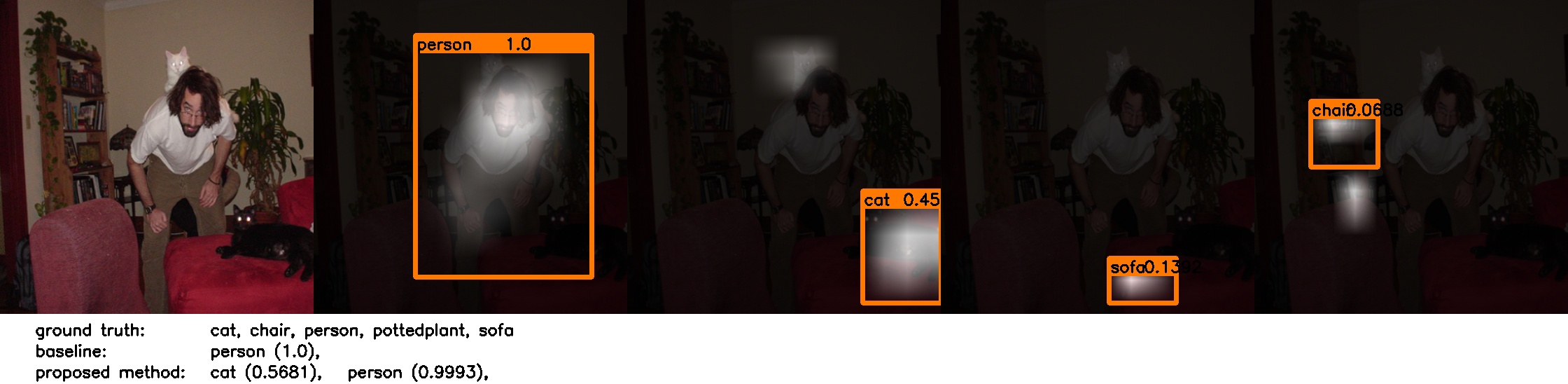}
		\vspace{-25pt}
		\begin{center}(d)\end{center}
		\vspace{5pt}
	\end{minipage}
	\caption{Visualization of results. 
		The first column shows the original input image, and the second to fifth columns show the visualisation of the five activation maps of the categories corresponding to the highest scores in the global branch, arranged in descending order.
		Take (b) as an illustration. In the global branch, the model can only detect the category of person, while the confidence scores for TV monitor is only 0.33, which is less than 0.5, leading to the multi-label classification result being significantly different compared to the ground-truth. However, after local cropping and feeding into the local branch for re-classification, the final confidence of the TV monitor improves from 0.33 to 0.53. We can observe that the re-classification of the model for the local region can effectively improve the multi-label classification performance.}
	\label{fig:vis}
\end{figure*}

\textbf{Feature Maps Chosen for $\bm{F_{r}}, \bm{F_{w}}$.} 
In the global branch, the augmented samples are fed into ResNet101 \cite{ResNet} for visual feature extraction, extracted feature maps with resolutions $1/8$, $1/16$ and $1/32$, denoted by $\bm{F_{8}}$, $\bm{F_{16}}$ and $\bm{F_{32}}$, respectively.
As for the choice of $\bm{F_{w}}$, since $\bm{F_{32}}$ contains more high-level semantic information, which is more conducive to the layer-by-layer mapping and gradient back-propagation of our category-aware weak supervision, we conjecture that $\bm{F_{w}}$ should be chosen as $\bm{F_{32}}$ to obtain the best performance. Regarding the choice of $\bm{F_{r}}$, we refer to the common practice in object detection methods \cite{ren2015faster}, as a result, $\bm{F_{16}}$ can be the only choice to obtain the best performance.
In order to analyze what size of the feature map should be selected for $\bm{F_{w}}, \bm{F_{r}}$,
we assign $\bm{F_{8}}, \bm{F_{16}}, \bm{F_{32}}$ to $\bm{F_{r}}$, and also assign $\bm{F_{8}}, \bm{F_{16}}, \bm{F_{32}}$ to $\bm{F_{w}}$, respectively. Then, the parameters of the input sizes of the convolutional and fully connected layers in the network structure are changed simultaneously to train the network and obtain the mAP values of the final experimental results. According to Tab. \ref{tab:feature_roipooling}, choosing $\bm{F_{16}}$ as $\bm{F_{r}}$ and choosing $\bm{F_{32}}$ as $\bm{F_{w}}$ will produce the best performance, which not only proves our conjecture but also further points out the validity and necessity of the proposed pipeline structure.

\begin{table}[ht]
	\renewcommand\arraystretch{\AblationTableSpacing}
	\centering
	\tablecaptionskip
	\caption{Ablation study on the impact of the chosen for $\bm{F_{r}}, \bm{F_{w}}$ on the performance. Experiments are conducted on the VOC dataset.}
	\AblationTableFontSize
	\setlength\tabcolsep{6.5pt}
	\begin{tabular}{c|ccc}
		\hline
		\textbf{$\bm{F_{r}}, \bm{F_{w}}$} & \textbf{$\bm{F_{w}}\leftarrow \bm{F_{8}}$} & \textbf{$\bm{F_{w}}\leftarrow  \bm{F_{16}}$} & \textbf{$\bm{F_{w}}\leftarrow \bm{F_{32}}$} \\
		\hline \hline
		\textbf{$\bm{F_{r}}\leftarrow  \bm{F_{8}}$} & 92.89 & 92.01  & 92.81  \\
		\textbf{$\bm{F_{r}}\leftarrow  \bm{F_{16}}$} & 91.00 & 93.27  & \textbf{95.17}  \\
		\textbf{$\bm{F_{r}}\leftarrow  \bm{F_{32}}$} & 93.02 & 93.46  & 94.00  \\ \hline
	\end{tabular}%
	\label{tab:feature_roipooling}%
\end{table}%

\textbf{Visualization of Results.}
Figure \ref{fig:vis} illustrates how the model can produce performance improvements. On the left is the original image, and the four images on the right are cropped areas used for local re-classification. These four crops were sent to the proposed local branch, where they were re-classified and made a correction for our final results. The strategy allows the model to concentrate on smaller objects and complex samples than the baseline and obtain higher classification performance, which further illustrates the effectiveness of the proposed method in this paper.

Despite the impressive performance improvements of our method, the computational cost (MACs) increases by only 14\% compared to the baseline \cite{ResNet}.
In addition, to further analyze the effectiveness of each module, we conducted extensive experiments,
including experiments on comparison of energy-based strategy versus area-size-based strategy, the specific descriptions and discussions of which are presented in supplementary material B.

\sectionvspace
\section{Conclusion}
In this work, we propose a novel multi-label image classification method to alleviate the limitations of previous region-based approaches.
Specifically, this paper proposes category-aware weak supervision as well as energy-based ROI selection to generate region proposals.
Moreover, by using cross-granularity attention to explore the correlation between global and local, label dependencies can be established implicitly.
Experimental results on two benchmark datasets demonstrate that our framework achieves superior performance over state-of-the-art methods.
In addition, the effectiveness of each component was also carefully studied.
\begin{acks}
	This work was supported by the National Key Research and Development Program of China (2021ZD0111000), 
	National Natural Science Foundation of China (62176092), 
	Shanghai Science and Technology Commission (21511100700), 
	Natural Science Foundation of Shanghai (20ZR1417700).
\end{acks}
\vfill\eject
\bibliographystyle{ACM-Reference-Format}
\balance
\bibliography{sample-base}

\appendix
\clearpage
\section*{Supplementary Material}

\section{Additional descriptions}
\begin{figure}[h]
    \centering
    \includegraphics[width=\linewidth]{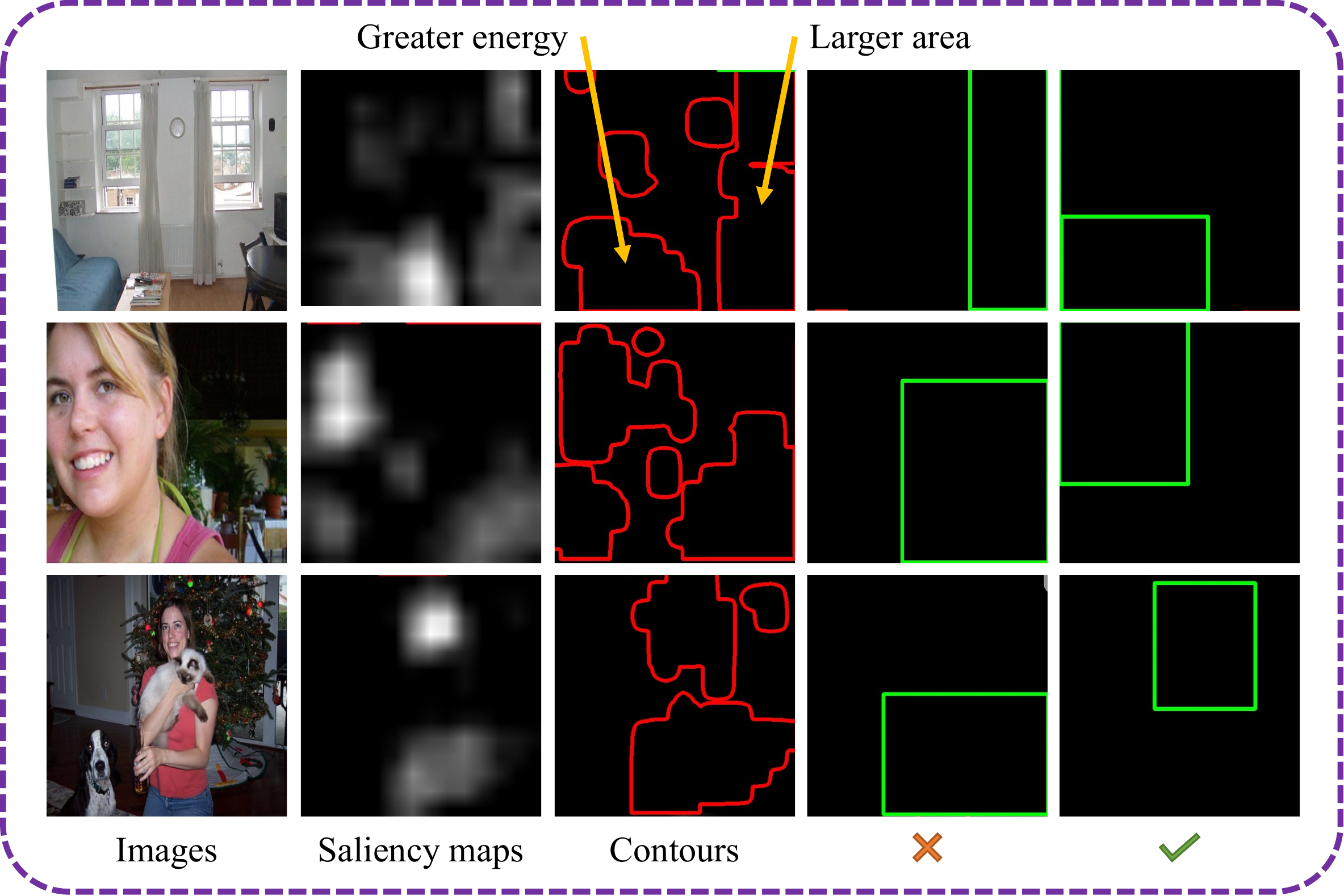}
    \caption{The strategy of Energy-based ROI generation. The first column on the left is the original input image, the second column is the saliency maps, the third column is the contours, the fourth is the sub-optimal selection strategy (based on area size) and the fifth column is the optimal selection strategy (based on energy).}
    \label{fig:energy}
\end{figure}

\textbf{The strategy of Energy-based ROI generation.}
Examples of our energy-based ROI selection are shown in Figure \ref{fig:energy}.
It can be demonstrated that more optimal regions can be selected by energy-based selection than by area-size-based selection.

\begin{figure}[h]
    \centering
    \includegraphics[width=\linewidth]{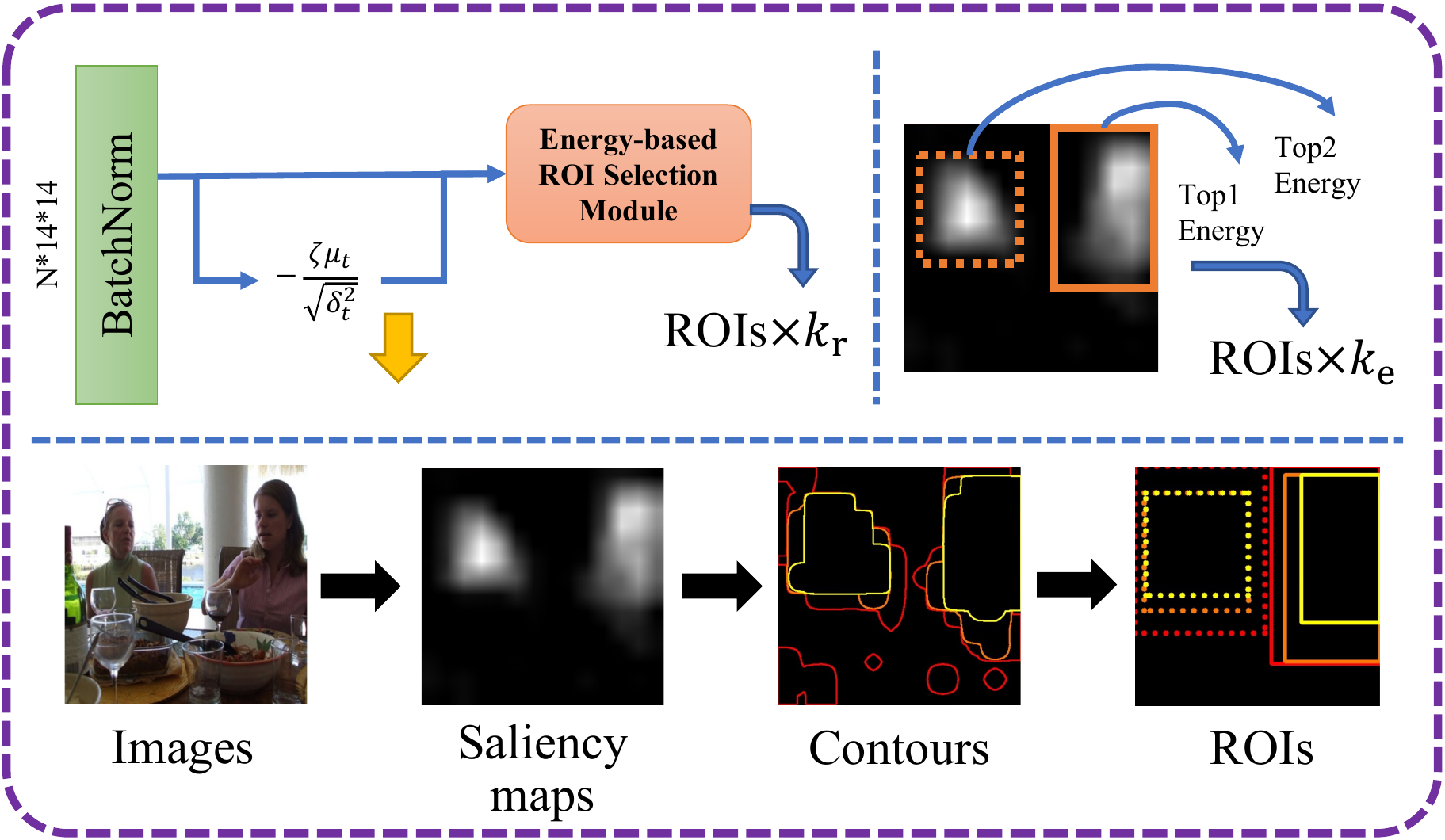}
    \caption{The ROI selection strategy when dealing with small objects. The {\color{red}\textbf{red}}, {\color{orange}\textbf{orange}} and {\color{yellow}\textbf{yellow}} rectangles represent the different proposals obtained by scaling strategy in the same region, while the \textbf{solid line} represents the candidate with the highest energy value and the \textbf{dashed line} represents the candidate with the second highest energy value.}
    \label{fig:coco}
\end{figure}

\begin{figure*}[!h]
  \centering
  \includegraphics[width=\linewidth]{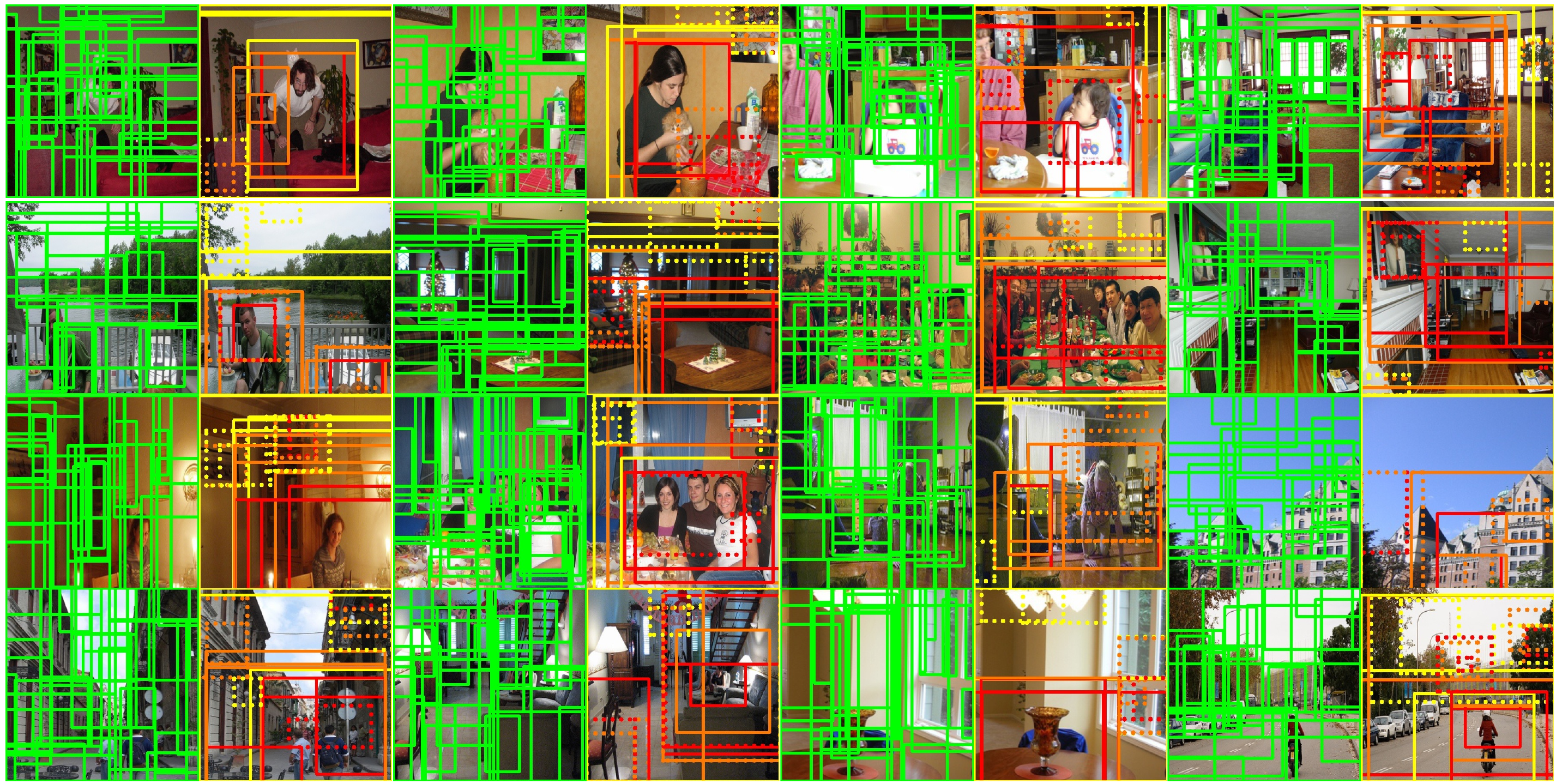}
  \caption{Comparison between our proposed ROI selection method and object detection technique,
  selective search 
  % \cite{uijlings2013selective}.
  The {\color{green}\textbf{green}} bounding boxes on the left are the proposals generated by the selective search,
  while the bounding boxes on the right are the proposals produced by our method,
  where {\color{red}\textbf{red}},
  {\color{orange}\textbf{orange}} and {\color{yellow}\textbf{yellow}} bounding boxes represent the different proposals obtained by scaling strategy in the same region.
  Besides, the \textbf{solid line} represents the option with the highest energy value and the \textbf{dashed line} represents the option with the second highest energy value.}
  \label{fig:vis_proposal}
\end{figure*}

To ensure that more semantic information about small objects is available for local branches since MS-COCO contains a large number of small objects \cite{lin2014microsoft,kisantal2019augmentation}, 
more regions are needed to improve the coverage of small areas.
Scaling to obtain more bounding boxes of smaller size can solve our needs.
Since we applied the $ReLU$ activation function (in the front of the energy-based ROI selection) to filter the negative activation values and use the regions with positive activation values as candidate regions, we can change the size of the bounding box to obtain more bounding boxes by adjusting each pixel value in the cropped feature map, instead of scaling the size of bounding boxes directly.
In addition, to scale the size of bounding boxes directly, manual specifying thresholds for each separate dataset is inevitable, which may reduce the model's generalizability and also be very labor-consuming and computationally demanding.

Specifically, our proposed scaling method is adaptive to the selected feature maps.
First, for each mini-batch, Batch Normalization performs a scaling operation on the input features.
\begin{equation}
    \widehat{x_i}=\frac{x_i-\mu_B}{\sqrt{\delta_B^2+\epsilon}},
\end{equation}
where $\mu_B=\frac{1}{m}\sum_{i}^{m}x_i$, $ \delta_B^2=\frac{1}{m}\sum_{i}^{m}\left(x_i-\mu_B\right)^2$.
To shift the features after scaling, we reduce the running mean $\mu_t$ to obtain a feature map with a smaller connected region. To ensure that the final output is on the same scale, we also divide the value of the offset by the running variance $\sqrt{\delta_t^2}$. Finally, the extra output features became,
\begin{equation}
    \widehat{x_i}=\frac{x_i-\mu_B}{\sqrt{\delta_B^2+\epsilon}}-\frac{{\zeta\mu}_t}{\sqrt{\delta_t^2}},
\end{equation}
where $\zeta$ is the empirical value we use to adjust the magnitude of the offset.

Since we consider the connected areas of positive values in the feature map as indicators for obtaining the contours, when the number of positive pixels within the feature map decreases, the size of the corresponding bounding box decreases accordingly. Then, we find the minimum enclosing rectangles for each contour and save them as bounding boxes. 
In this way, $k_r$ pooled feature maps are obtained, where $(kr-1)$ is generated by our method and the other $1$ is the feature map of the original features after pooling.
In our experiments, $\zeta$ will be given two different values to obtain $k_r$=3 bounding boxes in each high energy region.

In summary, for each image, we pick the top $k_s$ feature maps of categories with the highest confidence score in the global branch; for each feature map, we choose the top $k_e$ connected region with the highest energy value; for each connected region, we perform scaling to obtain the $k_r$ bounding box, as shown in the Figure \ref{fig:coco}.
In the end, we can acquire $k_o=k_s\times k_r\times k_e$ number of candidate regions.

\section{Additional experiments}
To further analyze the effectiveness of each module, we conducted extensive experiments.

\textbf{Energy-based ROI selection vs. Area-size-based ROI selection.}
To further illustrate the necessity of our proposed strategy of energy-based ROI selection, which has been explained in the previous analysis, we conducted ablation experiments on the VOC dataset.
The experiments compared energy-based ROI selection (the region selection strategy based on energy value) with area-size-based ROI selection (the region selection strategy based on area size, i.e., a preference for large sized bounding boxes), and the results are shown in the Table \ref{tab:roi_strategy}.

\begin{table}[!h]
  \centering
\caption{Ablation study on the impact of the chosen for Energy-based ROI selection vs. Area-size-based ROI selection on the performance. Experiments are conducted on both VOC and COCO datasets.}

  \renewcommand\arraystretch{1.2}
  \begin{tabular}{c|cc}
      \hline
      \textbf{Methods} & \textbf{Energy-based} & \textbf{Area-size-based} \\
      \hline \hline
       \textbf{VOC} & 95.17 & 95.00 \\ \hline
   \textbf{COCO} & 85.31 & 84.23 \\ \hline
  \end{tabular}%
  \label{tab:roi_strategy}%
\end{table}%

\begin{figure*}[!h]
  \centering
  \vspace{5pt}
  \begin{minipage}[b]{1\textwidth}
    \includegraphics[width=\linewidth]{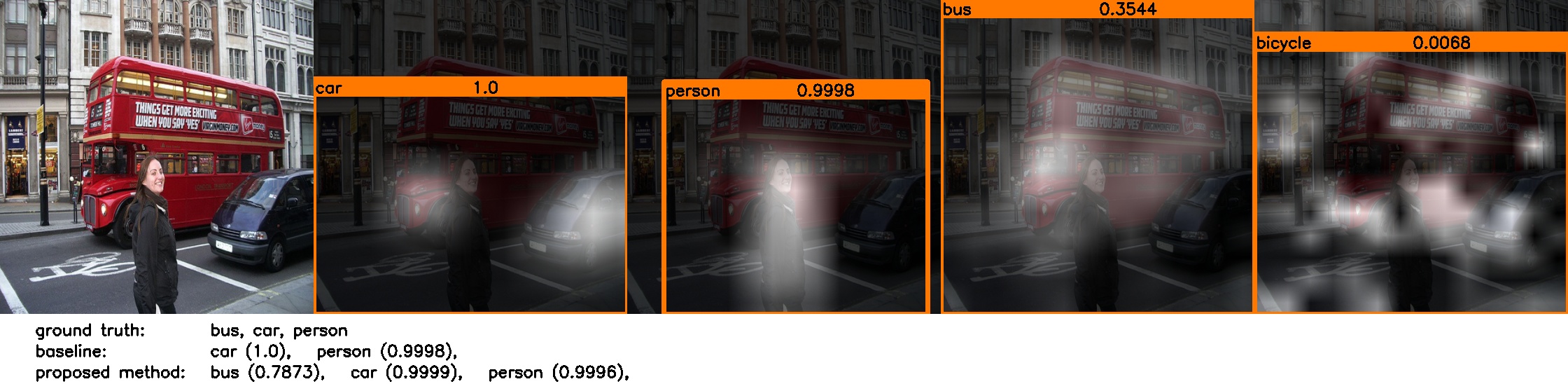}
    \vspace{-25pt}
    \begin{center}(a)\end{center}
    \vspace{5pt}
  \end{minipage}
  \begin{minipage}[b]{1\textwidth}
    \includegraphics[width=\linewidth]{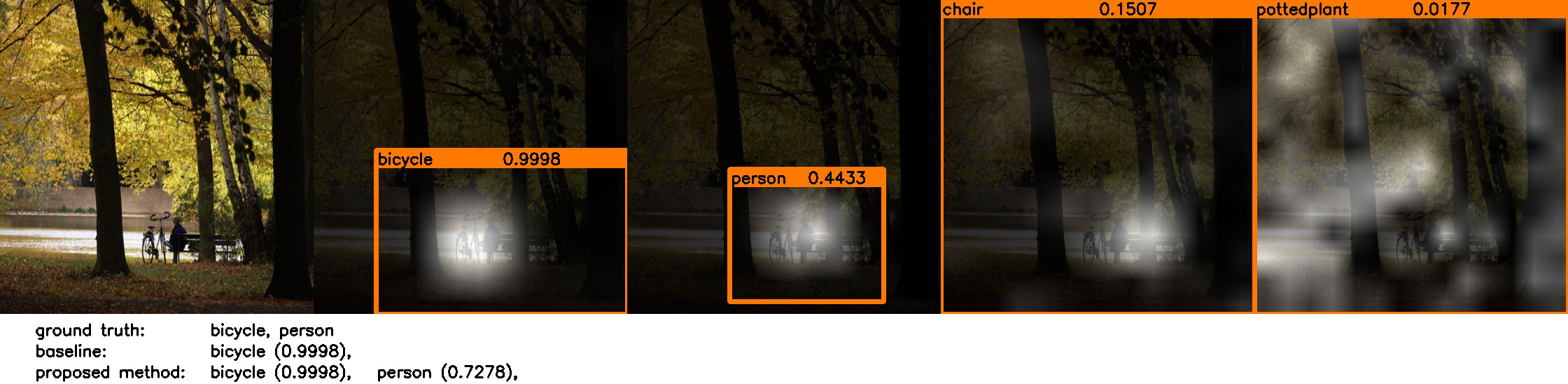}
    \vspace{-25pt}
    \begin{center}(b)\end{center}
    \vspace{5pt}
  \end{minipage}
  \begin{minipage}[b]{1\textwidth}
    \includegraphics[width=\linewidth]{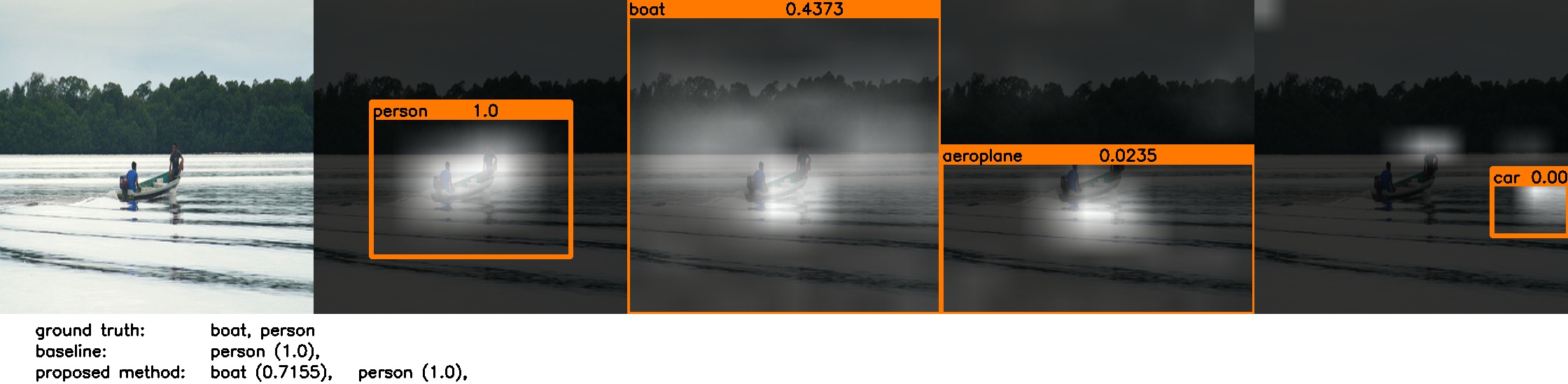}
    \vspace{-25pt}
    \begin{center}(c)\end{center}
  \end{minipage}
  \caption{Visualization of results. 
    The first column shows the original input image, and the second to fifth columns show the visualization of the five activation maps of the categories corresponding to the highest scores in the global branch, arranged in descending order.
    Although the recognition is relatively accurate, some bounding boxes are too large, which affects the efficiency of the methods.
    }
  \label{fig:fail}
\end{figure*}

As can be seen in Table \ref{tab:roi_strategy}, the classification performance of the energy-based ROI selection strategy is still improved over that of the area-based strategy, although the difference is not significant. Since we want to obtain the best performance in our experiments, we decided to use the energy-based ROI selection strategy.

\textbf{Computational Costs.} 
To illustrate the efficiency and computational costs of our model, we counted the number of multiply-accumulate (MAC) of resnet101, the model without attention (Our w/o CA\&GA), and the full model.
\begin{table}[!h]
    \centering
	\caption{Experiments on computational cost and efficiency analysis of the proposed model. 
    The experiments are based on 448-size image input to count the number of parameters and computational cost of each comparison model during inference.
    mAP values are obtained by inference on the MS-COCO 2014 dataset.}

    \renewcommand\arraystretch{1.2}
    \begin{tabular}{l|ccc}
        \hline
        \textbf{Methods} & \textbf{MACs(G)} & \textbf{Params(M)} & \textbf{mAP} \\
        \hline \hline
         \textbf{ResNet101} & 31.39 & 44.55 & 79.31 \\ \hline
         \textbf{Our w/o CA\&GA} & 32.43 & 57.63 & 81.67 \\ \hline
         \textbf{Full model} & 35.98 & 74.42 & 85.27 \\ \hline
    \end{tabular}%
    \label{tab:costs}%
\end{table}%

\begin{figure}[h]
    \begin{tikzpicture}
    \begin{axis}[
        width=8.0cm,height=5.0cm,
        xlabel=Input image size for inference process,
        ylabel=MACs(G),
        ylabel near ticks,
        tick align=inside,
        grid style=dashed,
        legend pos=north west,
        ymajorgrids=true,    
        xmajorgrids=true,    
        legend cell align={left},
        legend style={font=\footnotesize}
        ]
    \addplot[smooth,mark=triangle,red] plot coordinates { 
        (224,7.85)
        (448,31.39)
        (576,51.89)
        (640,64.06)
    };
    \addlegendentry{Resnet101}
    \addplot[smooth,mark=square,black] plot coordinates {
        (224,8.88)
        (448,32.43)
        (576,52.94)
        (640,65.12)
    };
    \addlegendentry{Our w/o CA\&GA}
        \addplot[smooth,mark=*,blue] plot coordinates {
            (224,9.80)
            (448,35.98)
            (576,58.91)
            (640,72.59)
        };
        \addlegendentry{Full model}
    \end{axis}
    \end{tikzpicture}
    \caption{Experiment on the computational cost of the model. 
    The experiments are based on different sizes of image input and are used to compare the effect of image size on the computational cost.
    }
    \label{fig:size_line}
    \end{figure}
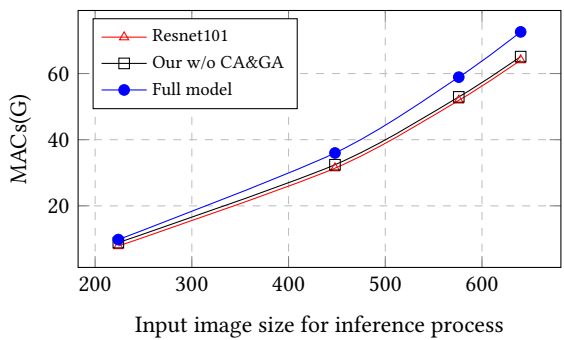

As can be seen from Table \ref{tab:costs}, there is only a slight increase in the computational cost ($\approx 14\%$), despite the acceptable increase in the number of parameters of the model (mainly on the cross-granularity attention), which further illustrates the great efficiency of our model.
This small increase in computation greatly improves the performance, making the model more practical.
Additionally, we analyze the growth of the corresponding computational cost of varying the input sizes. 
As can be seen from Figure \ref{fig:size_line}, the additional increase in computational cost of our model is small in proportion compared to the increase in the backbone computation due to the resolution change in input image.

\textbf{Visualization of Region Proposals.}
To demonstrate the advantage of our method in generating region proposals, Figure \ref{fig:vis_proposal} gives a visual comparison between the proposals generated by our method and that generated by selective search \cite{uijlings2013selective}. It is manifest from the Figure \ref{fig:vis_proposal} that the proposals generated by our method are more accurate and efficient (less number is needed) compared to selective search when locating possible objects in a given image, which illustrates the effectiveness of category-aware weak supervision and energy-based ROI selection.
Within the visualization results (Figure \ref{fig:fail}), we discovered several examples that could be limitations of our methods. Since the weakly supervised loss only constrains the non-existence/absence category and the threshold selection for the foreground is automatic, the region proposal's bounding box may be slightly larger. Although it is not a major issue, large local regions can occasionally interfere with the reclassification of small objects and influence the final fusion results, which can be an area that future work may address.

% \bibliographystyle{ACM-Reference-Format}
% \balance

\end{document}